\pdfoutput=1

\documentclass[11pt]{article}

\usepackage[final]{acl}

\usepackage{times}
\usepackage{latexsym}
\usepackage{graphicx}
\usepackage{amsmath}
\usepackage{amssymb}
\usepackage{multirow}
\usepackage{booktabs}
\usepackage{algorithm}
\usepackage{algorithmic}
\usepackage{hyperref}
\usepackage[dvipsnames]{xcolor}
\usepackage{cite}
\usepackage{subfigure}
\usepackage{array}
\usepackage{microtype}
\usepackage{makecell}
\usepackage{url}
\usepackage{tabularx}
\usepackage{colortbl}
\usepackage{float}
\usepackage{tipa}
\usepackage{placeins}
\usepackage{comment}
\usepackage{pifont}
\usepackage{tikz}
\usetikzlibrary{shapes,arrows,positioning,calc,patterns,decorations.pathreplacing}

\definecolor{mypink}{rgb}{0.9686274,0.882352,0.86666}
\definecolor{mygreen}{rgb}{0.819607,0.890196,0.760784}
\definecolor{headercolor}{RGB}{240,240,240}

\newcolumntype{L}[1]{>{\raggedright\let\newline\\\arraybackslash\hspace{0pt}}m{#1}}
\newcolumntype{C}[1]{>{\centering\let\newline\\\arraybackslash\hspace{0pt}}m{#1}}

\newcommand{\dataset}{\textsc{WritingPrompts}}

\usepackage[T1]{fontenc}
\usepackage[utf8]{inputenc}
\usepackage{microtype}
\usepackage{inconsolata}
\usepackage{graphicx}

\title{The Geometry of Creative Variability: \\How Credal Sets Expose Calibration Gaps in Language Models}

\author{
  \textbf{Esteban Garces Arias\textsuperscript{1,2}},
  \textbf{Julian Rodemann\textsuperscript{1,3}},
  \textbf{Christian Heumann\textsuperscript{1}}
\\
\\
  \textsuperscript{1}Department of Statistics, LMU Munich, Germany\\
  \textsuperscript{2}Munich Center for Machine Learning (MCML)\\
  \textsuperscript{3}CISPA Helmholtz Center for Information Security, Saarbrücken, Germany
\\ 
  \small{
    \textbf{Correspondence:} \href{mailto:Esteban.GarcesArias@stat.uni-muenchen.de}{Esteban.GarcesArias@stat.uni-muenchen.de}
  }
}

\begin{document}
\maketitle
\begin{abstract}
Understanding uncertainty in large language models remains a fundamental challenge, particularly in creative tasks where multiple valid outputs exist. We present a geometric framework using credal sets—convex hulls of probability distributions—to quantify and decompose uncertainty in neural text generation, calibrated against human creative variation. Analyzing 500 creative writing prompts from the \dataset{} dataset with 10 unique human continuations each, we evaluate four language models across five decoding strategies, generating 100,000 stories. Our credal set analysis reveals substantial gaps in capturing human creative variation, with the best model-human calibration reaching only 0.434 (Gemma-2B with temperature 0.7). We decompose total uncertainty into \textit{epistemic} and \textit{aleatoric} components, finding that the choice of decoding strategy contributes 39.4\% to 72.0\% of total epistemic uncertainty. Model scale shows weak correlation with calibration quality and no significant difference exists between base and instruction-tuned models in calibration quality. Our geometric framework provides actionable insights for improving generation systems for human-AI creative alignment. We release our complete experimental framework at \url{https://github.com/EstebanGarces/uncertainHuman}.
\end{abstract}

\section{Introduction}

\begin{figure}[t]
\centering
\begin{tikzpicture}[scale=1.0]

\definecolor{human}{RGB}{46,134,171}
\definecolor{base}{RGB}{162,59,114}  
\definecolor{inst}{RGB}{241,143,1}

\node[draw=gray!30, fill=gray!5, text width=\columnwidth-10pt, inner sep=6pt, font=\footnotesize] (textbox) at (0,0) {
\textbf{Prompt:} \textit{``The last person on Earth sits alone. There is a knock on the door.''}
\vspace{3pt}

\textbf{Human continuations:}\\
\textcolor{human}{$\bullet$} ``My heart stopped. After three years of silence...''\\
\textcolor{human}{$\bullet$} ``I laughed. The universe's final joke...''\\
\textcolor{human}{$\bullet$} ``Pizza delivery,' a voice called out...''
\vspace{3pt}

\textbf{Model continuations (Instruct):}\\
\textcolor{inst}{$\bullet$} ``The survivor cautiously approached the door...''\\
\textcolor{inst}{$\bullet$} ``They slowly walked to the door, heart pounding...''\\
\textcolor{inst}{$\bullet$} ``With trembling hands, the survivor reached...''
};

\begin{scope}[yshift=-4.2cm]

\draw[gray!30, fill=white] (-3.2,-1.6) rectangle (3.2,1.6);

\draw[->, gray!70] (-2.8,0) -- (2.8,0) node[below, font=\scriptsize] at (1.4,-0.1) {Semantic 1};
\draw[->, gray!70] (0,-1.3) -- (0,1.3) node[right, font=\scriptsize] at (0.1,1.0) {Semantic 2};

\draw[gray!10, very thin] (-2.8,-1.3) grid[step=0.5] (2.8,1.3);

\draw[human, line width=1.5pt, dashed, fill=human!5] (0,0) ellipse (2.3cm and 1.0cm);
\draw[base, line width=1.5pt, dashed, fill=base!5] (0.2,0) ellipse (1.4cm and 0.6cm);
\draw[inst, line width=1.5pt, dashed, fill=inst!5] (0.3,0) ellipse (0.5cm and 0.25cm);

\foreach \p in {(-2.0,0.6),(-1.5,-0.7),(-0.8,0.8),(0.2,-0.6),
                (0.9,0.7),(1.6,-0.4),(2.0,0.3)} 
    \fill[human] \p circle (1.8pt);

\foreach \p in {(-0.6,0.3),(0.1,-0.3),(0.6,0.4),(1.0,-0.2),(1.2,0.2)} 
    \fill[base] \p circle (1.8pt);

\foreach \p in {(0.25,0.08),(0.35,-0.05),(0.3,0.12),(0.4,-0.02)} 
    \fill[inst] \p circle (1.8pt);

\node[human, font=\scriptsize\bfseries] at (-1.5,0.7) {Human};
\node[base, font=\scriptsize\bfseries] at (1.0,0.4) {Base};
\node[inst, font=\scriptsize\bfseries] at (0.3,-0.4) {Instruct};

\node[font=\scriptsize] at (0,-1.9) {
\textcolor{human}{$\bullet$ Human (n=5000)} \quad 
\textcolor{base}{$\bullet$ Base model} \quad 
\textcolor{inst}{$\bullet$ Instruction-tuned}
};

\end{scope}

\end{tikzpicture}
\caption{Examples of human versus model creative variation. \textbf{Top:} Continuations show diverse human interpretations versus convergent model responses. \textbf{Bottom:} Credal sets (dashed ellipses) represent convex hulls of diversity distributions in semantic, lexical, and syntactic space.}
\label{fig:human_vs_model}
\end{figure}
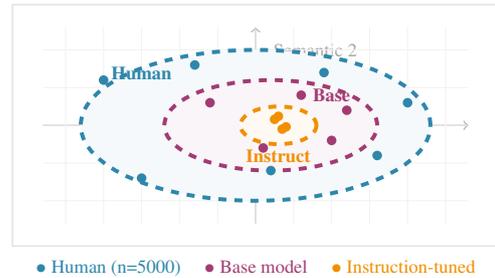

The deployment of large language models in creative and open-ended applications demands not merely generating plausible text, but understanding and calibrating the uncertainty inherent in these generations. While uncertainty quantification has been extensively studied in discriminative tasks \citep{gal2016dropout, lakshminarayanan2017simple, Ovadia2019CanYT}, the challenge becomes substantially more complex in generative settings where no single ground truth exists and quality itself becomes a multidimensional construct \citep{garces-arias-etal-2025-towards, arias2025statisticalmulticriteriaevaluationllmgenerated}. This complexity is particularly acute in creative writing, where the same prompt can inspire substantially different narratives, styles, and interpretations (cf. Figure \ref{fig:human_vs_model}).

\noindent Current approaches to uncertainty quantification in language models predominantly focus on token-level probabilities or computationally expensive ensemble methods \citep{ling2024uncertainty, zhang2025token}. These methods, while valuable, fail to capture the semantic, lexical, and syntactic-level uncertainty that determines whether a model appropriately captures the breadth of human creative expression. More fundamentally, existing frameworks lack principled methods for distinguishing between \textit{aleatoric uncertainty}—the irreducible variation inherent in creative tasks—and \textit{epistemic uncertainty} arising from model limitations. This distinction proves crucial for both improving model design and establishing appropriate deployment boundaries. In this work, we address these limitations through a novel framework that leverages human variation as a natural calibration target for model uncertainty. Our key insight is that multiple human responses to the same creative prompt provide a direct empirical measure of aleatoric uncertainty. By representing both human and model variation as credal sets—convex hulls of probability distributions over textual characteristics—we can geometrically analyze whether models exhibit appropriate uncertainty: high variation when humans disagree, and convergent outputs when humans reach consensus. This credal set approach offers several theoretical and practical advantages over existing methods. Theoretically, it provides a rigorous framework for uncertainty decomposition that respects the inherently distributional nature of creative variation. Each prompt induces its own distribution over possible continuations, and the collection of these distributions across many prompts forms a credal set that fully characterizes the uncertainty landscape. Practically, this framework enables direct comparison between human and the model's uncertainty through geometric measures such as overlap coefficients, Hausdorff distance \citep{232073}, and volume ratios. Our empirical investigation analyzes 500 carefully selected prompts from the \dataset{} dataset, each accompanied by 10 verified unique human continuations totaling 5,000 human-written stories. We evaluate four language models—GPT2-XL \citep{radford2019language}, Gemma-2B \citep{gemmateam2024gemma2improvingopen}, Mistral-7B-Instruct-v0.2 \citep{jiang2023mistral}, and Llama-3.1-8B-Instruct \citep{dubey2024llama}—generating 10 samples per configuration across five decoding strategies, yielding 100,000 model-generated stories. Through comprehensive analysis of semantic, lexical, and syntactic diversity, we construct and compare credal sets that reveal systematic patterns in how models capture or fail to capture human-like variation.

\paragraph{Contributions}
\begin{itemize}
    \item We introduce credal sets—convex hulls of diversity distributions—as a geometric framework for quantifying uncertainty in open-ended text generation.
    \item We analyze 100,000 generated stories, finding that the best model-human calibration reaches only 0.434 (Gemma-2B with temperature 0.7), revealing substantial gaps in creative variation.
    \item We show weak correlation between model scale and calibration (Spearman's $\rho = 0.400$, $p = 0.600$) and no significant difference between base and instruction-tuned models ($t = -0.712$, $p = 0.486$).
    \item We decompose uncertainty to reveal that decoding strategy choice contributes 39.4-72.0\% of total epistemic uncertainty, with base models showing higher sensitivity.
    \item We release our complete experimental framework and datasets for reproducible research.\footnote{\url{https://github.com/EstebanGarces/uncertainHuman}}
\end{itemize}

\section{Related Work}

Uncertainty quantification in language models has emerged as a critical research area, particularly as these models are deployed in high-stakes applications. We organize our discussion around three main themes: theoretical frameworks for uncertainty decomposition, practical estimation methods, and uncertainty-aware generation strategies.

\subsection{Theoretical Frameworks for Uncertainty Decomposition}

The foundational challenge lies in decomposing total predictive uncertainty into meaningful components. \citet{ling2024uncertainty} address this for in-context learning scenarios: They derive total predictive uncertainty through the classical additive information-theoretic decomposition, where the first term captures aleatoric uncertainty (inherent randomness in the task) and the second represents epistemic uncertainty (model uncertainty). They propose entropy estimators based on variational bounds on mutual information for practical approximation. However, \citet{wimmer2023quantifying} note that this distinction can become ambiguous in pre-trained models where the training distribution itself is uncertain.

The use of credal sets for uncertainty representation has been common in classification tasks \citep{zaffalon2003tree}, but, to the best of our knowledge, our work is the first to apply this framework to open-ended generation. Credal sets provide a natural representation for situations where a single probability distribution is insufficient to capture uncertainty, instead maintaining a set of plausible distributions \citep{Levi1980-LEVTEO-7}.

\subsection{Practical Methods for Uncertainty Estimation}

Various practical approaches for uncertainty estimation have been recently proposed: \citet{lin2022teachingmodelsexpressuncertainty, xiong2024llmsexpressuncertaintyempirical} explore methods to verbalize uncertainty,  \citet{kadavath2022language, liu2024uncertaintyestimationquantificationllms, ulmer2024calibratinglargelanguagemodels} focus on probes for LLM calibration, while \citet{pitis2023boostedpromptensembleslarge, hou2024decomposinguncertaintylargelanguage} have focused on self-consistency approaches.

\noindent Recent work has developed various approaches to estimate uncertainty without expensive ensemble methods. \citet{zhang2025token} introduce a training-free method injecting low-rank random weight perturbations during decoding to estimate token-level uncertainties. These are aggregated into sequence-level measures that correlate strongly with correctness on mathematical reasoning benchmarks, with epistemic uncertainty effectively identifying incorrect reasoning paths. While this perturbation approach elegantly estimates model uncertainty, it focuses on uncertainty from a single fixed model. Our work examines uncertainty arising from different decoding strategies and model architectures, providing a complementary perspective on variation sources in language model outputs. \citet{yadkori2024believe} propose an information-theoretic metric based on mutual information over iteratively prompted responses, interpreting heavy dependencies between subsequent responses as indicators of high epistemic uncertainty and potential hallucination, though requiring computationally expensive multiple inference passes. \citet{aichberger2024rethinking} pursue efficiency with a single-pass approximation using negative log-likelihood of greedy outputs, proving that high NLL correlates with high epistemic uncertainty under certain assumptions.

\subsection{Uncertainty-Aware Generation and Human Baselines}

\citet{garces-arias-etal-2024-adaptive, ding-etal-2025-guard} propose uncertainty-aware decoding that dynamically adjusts generation parameters based on local uncertainty. They compute entropy $H(p_t)$ of the token probability distribution $p_t$ at each generation step $t$ and adjust the truncation threshold dynamically, demonstrating that uncertainty signals can improve generation quality in real-time. Most directly related to our work, \citet{giulianelli-etal-2023-comes} evaluate uncertainty in neural text generators against human production variability, arguing that well-calibrated models should exhibit similar variation to humans. They analyze GPT-2 on story generation with limited prompts, finding that it under-produces diversity relative to human baselines. Our work substantially extends this research by: (1) scaling to 500 prompts with 10 unique continuations each, (2) including contemporary instruction-tuned models, (3) evaluating five decoding strategies systematically, (4) explicitly decomposing uncertainty into aleatoric and epistemic components, and (5) providing quantitative calibration metrics based on credal set overlap coefficients.

\section{Methodology}

\subsection{Dataset Construction and Human Baselines}

The \dataset{} dataset \citep{fan2018hierarchical} provides naturalistic creative writing data from Reddit's r/WritingPrompts community. We implement rigorous selection criteria to ensure data quality:

\begin{enumerate}
    \item \textbf{Uniqueness verification}: We compute MD5 hashes for all stories and select only prompts with exactly 10 unique continuations, eliminating duplicates that could bias diversity measurements.
    \item \textbf{Length filtering}: We retain prompts between 20-500 characters and stories between 52-987 tokens (mean: 312.4, std: 148.2), ensuring sufficient content for meaningful analysis while avoiding outliers.
    \item \textbf{Quality scoring}: We prioritize prompts by the diversity of story lengths they elicit (measured by standard deviation), selecting those that inspire varied responses rather than formulaic continuations.
\end{enumerate}

\noindent This process yields 500 high-quality prompts with 5,000 unique human stories, providing a robust baseline for calibration analysis.

\subsection{Model Selection and Configuration}

Our model selection explores the calibration landscape across different architectures and training paradigms:

\paragraph{Base models:} GPT2-XL (1.5B) \citep{radford2019language} serves as a canonical autoregressive baseline, while Gemma-2B \citep{gemmateam2024gemma2improvingopen} represents modern architectural improvements at comparable scale. These models, trained on diverse internet text without explicit instruction following, potentially preserve more natural variation patterns.

\paragraph{Instruction-tuned models:} Mistral-7B-Instruct-v0.2 \citep{jiang2023mistral} and Llama-3.1-8B-Instruct \citep{dubey2024llama} represent strong open-source models with instruction tuning and alignment. While offering improved controllability, we investigate whether alignment training constrains creative exploration\footnote{All models use appropriate prompt formatting with careful post-processing to remove prompt artifacts from generations, ensuring fair comparison across architectures.}.

\subsection{Decoding Strategy Design}

We systematically evaluate five decoding strategies that control output diversity through different mechanisms:

\begin{itemize}
    \item \textbf{Temperature scaling} ($\tau \in \{0.7, 1.2\}$): Directly modulates the entropy of the output distribution \citep{ackley1985learning}
    \item \textbf{Nucleus sampling} ($p=0.9$): Dynamically adjusts the token consideration set based on cumulative probability \citep{holtzman2020curious}
    \item \textbf{Top-$k$ sampling} ($k=40$): Maintains a fixed-size token pool \citep{fan2018hierarchical}
    \item \textbf{Typical sampling} ($p=0.95$): Selects tokens based on expected information content \citep{meister-etal-2023-locally}
\end{itemize}

\noindent Each configuration generates 10 independent samples with different random seeds, totaling 100,000 model-generated stories for analysis.

\subsection{Diversity Metrics}

Our metric suite captures multiple dimensions of textual variation through pairwise distance-based measures following \citet{giulianelli-etal-2023-comes}:

\subsubsection{Semantic Diversity}

We compute semantic diversity as the mean pairwise cosine distance between Sentence-BERT embeddings \citep{reimers2019sentence}:

$$D_{\text{sem}}(\mathcal{S}) = \frac{2}{|\mathcal{S}|(|\mathcal{S}|-1)} \sum_{i<j} (1 - \cos(e_i, e_j)),$$

\noindent where $e_i$ represents the embedding of story $i$ using the all-MiniLM-L6-v2 model \citep{wang2020minilmdeepselfattentiondistillation}. This captures high-level narrative and thematic variation.

\subsubsection{Lexical Diversity}

We measure lexical diversity using Jaccard distance between word unigrams:

$$D_{\text{lex}}(\mathcal{S}) = \frac{2}{|\mathcal{S}|(|\mathcal{S}|-1)} \sum_{i<j} \left(1 - \frac{|V_i \cap V_j|}{|V_i \cup V_j|}\right),$$

\noindent where $V_i$ represents the vocabulary set of story $i$. This captures variation in word choice and vocabulary richness.

\subsubsection{Syntactic Diversity}

We measure syntactic variation through Jaccard distance of part-of-speech (POS) bigrams:

$$D_{\text{syn}}(\mathcal{S}) = \frac{2}{|\mathcal{S}|(|\mathcal{S}|-1)} \sum_{i<j} \left(1 - \frac{|P_i \cap P_j|}{|P_i \cup P_j|}\right),$$

\noindent where $P_i$ represents the set of POS bigrams extracted using spaCy's en\_core\_web\_sm model \citep{spacy2}. This captures stylistic and structural variation in the generated text.

\subsection{Theoretical Framework: Credal Sets}

Our methodology rests on the principle that uncertainty in creative text generation should be understood relative to the natural variation exhibited by humans facing the same creative task. We formalize this through a credal set framework that captures uncertainty as a set of plausible probability distributions rather than a single distribution.

For a given prompt $p$, let $\mathcal{H}_p = \{h_1, ..., h_{10}\}$ denote the set of human continuations and $\mathcal{M}_{p,m,d} = \{s_1, ..., s_{10}\}$ denote the set of model continuations for model $m$ using decoding strategy $d$. For any set of continuations $\mathcal{S}$, we compute a diversity vector $\mathbf{v}_p = [D_{\text{sem}}(\mathcal{S}), D_{\text{lex}}(\mathcal{S}), D_{\text{syn}}(\mathcal{S})]$.

The human credal set for a collection of prompts $\mathcal{P}$ is then defined as:
$$\mathcal{C}_H = \text{ConvexHull}\left(\{\mathbf{v}_p^H : p \in \mathcal{P}\}\right),$$

\noindent where each $\mathbf{v}_p^H$ is the diversity vector computed from human continuations for prompt $p$.

\noindent Similarly, the model credal set for a specific configuration $(m, d)$ is:
$$\mathcal{C}_{M,d} = \text{ConvexHull}\left(\{\mathbf{v}_p^{M,d} : p \in \mathcal{P}\}\right).$$

\noindent The convex hull is computed using the Quickhull algorithm \citep{10.1145/235815.235821} after standardizing the diversity vectors. This representation enables geometric analysis of uncertainty relationships through set operations and distance metrics.

\subsection{Calibration Analysis}
Calibration quality is assessed through the overlap coefficient of credal sets:
$$\text{Calibration}(M,d) = \text{Overlap}(\mathcal{C}_H, \mathcal{C}_{M,d}),$$
\noindent where overlap is computed using nearest-neighbor distances between credal set vertices. The overlap coefficient is calculated as:
\begin{align*}
\text{Overlap} = \frac{1}{2}\bigg(&\frac{|\{v \in V_M : d(v, V_H) < \theta\}|}{|V_M|} \notag\\
&+ \frac{|\{v \in V_H : d(v, V_M) < \theta\}|}{|V_H|}\bigg),
\end{align*}
\noindent where $V_M$ and $V_H$ are the vertex sets of the model and human credal sets respectively, $d(v, V)$ is the minimum distance from point $v$ to set $V$, and $\theta$ is an adaptive threshold set to half the mean variance scale. Values range from 0 (disjoint sets) to 1 (perfect overlap).

\subsection{Uncertainty Decomposition}

To decompose uncertainty, we leverage variation across decoding strategies. For a given model $M$, we collect all diversity vectors across different strategies and compute:

    \paragraph{Strategy centroids}:\\ $\mathbf{c}_d = \text{mean}(\{\mathbf{v}_p^{M,d} : p \in \mathcal{P}\})$ for each strategy $d$
    \paragraph{Between-strategy variance}:\\$\sigma^2_{\text{between}} = \text{Var}(\{\mathbf{c}_d : d \in \mathcal{D}\})$
    \paragraph{Within-strategy variance}:\\ $\sigma^2_{\text{within}} = \text{mean}_d[\text{Var}(\{\mathbf{v}_p^{M,d} : p \in \mathcal{P}\})]$

\noindent The epistemic ratio is then:
$$\text{Epistemic}_M = \frac{\sigma^2_{\text{between}}}{\sigma^2_{\text{between}} + \sigma^2_{\text{within}}}.$$

\noindent This quantifies the proportion of uncertainty arising from configuration choices rather than inherent task ambiguity.

\begin{figure*}[t]
\centering
\includegraphics[width=\linewidth]{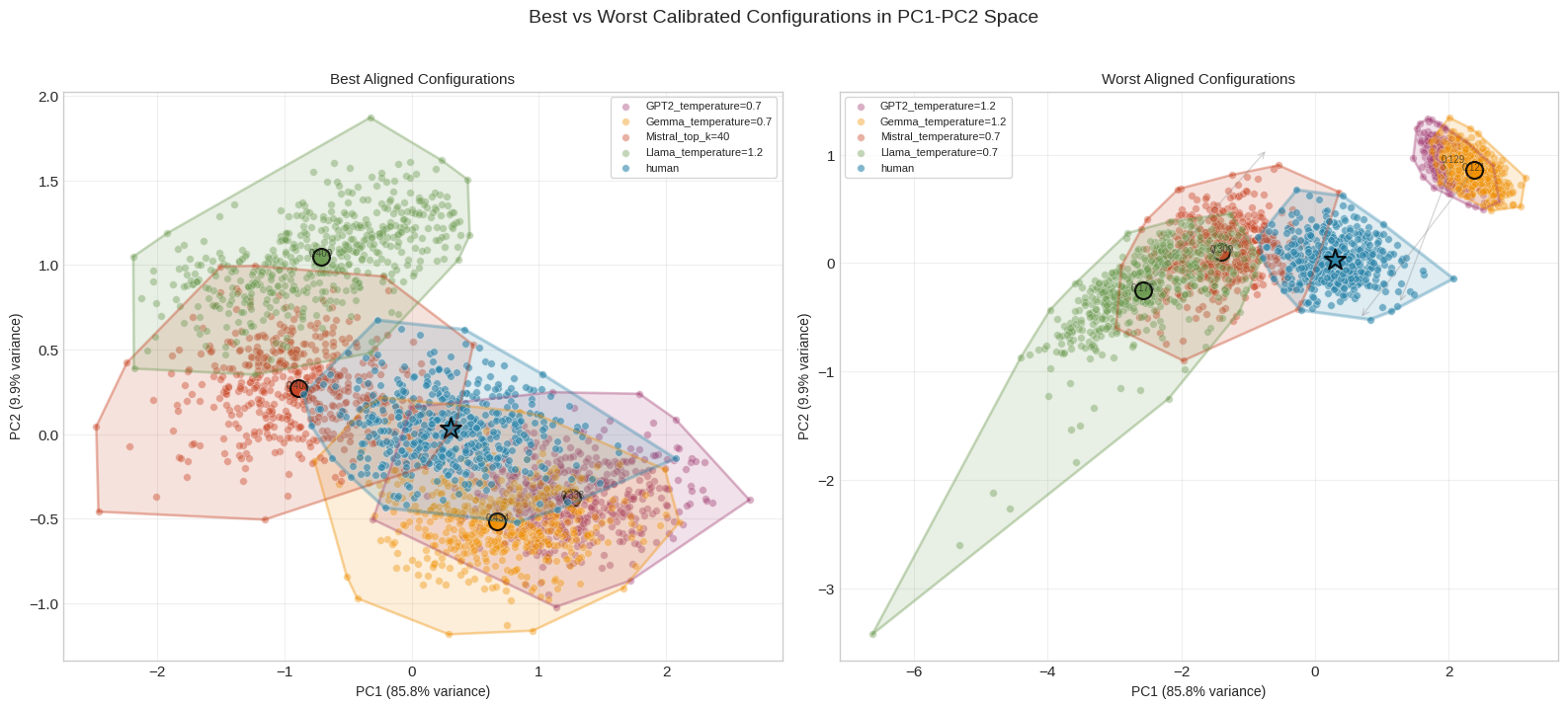}
\caption{Credal sets visualization in principal component space. Human creative variation (blue) and model-generated variation exhibit different geometric patterns and a high sensitivity with respect to the decoding configuration. Points represent diversity vectors from individual prompts; convex hulls indicate credal set boundaries. PC1 explains 85.8\% of the variance, suggesting a strong correlation between diversity dimensions. Best (left) and worst aligned configurations (right), measured by the overlap of the credal sets, are presented.}
\label{fig:credal_visualization}
\end{figure*}

\section{Results}

\begin{figure}[t]
\centering
\includegraphics[width=\columnwidth]{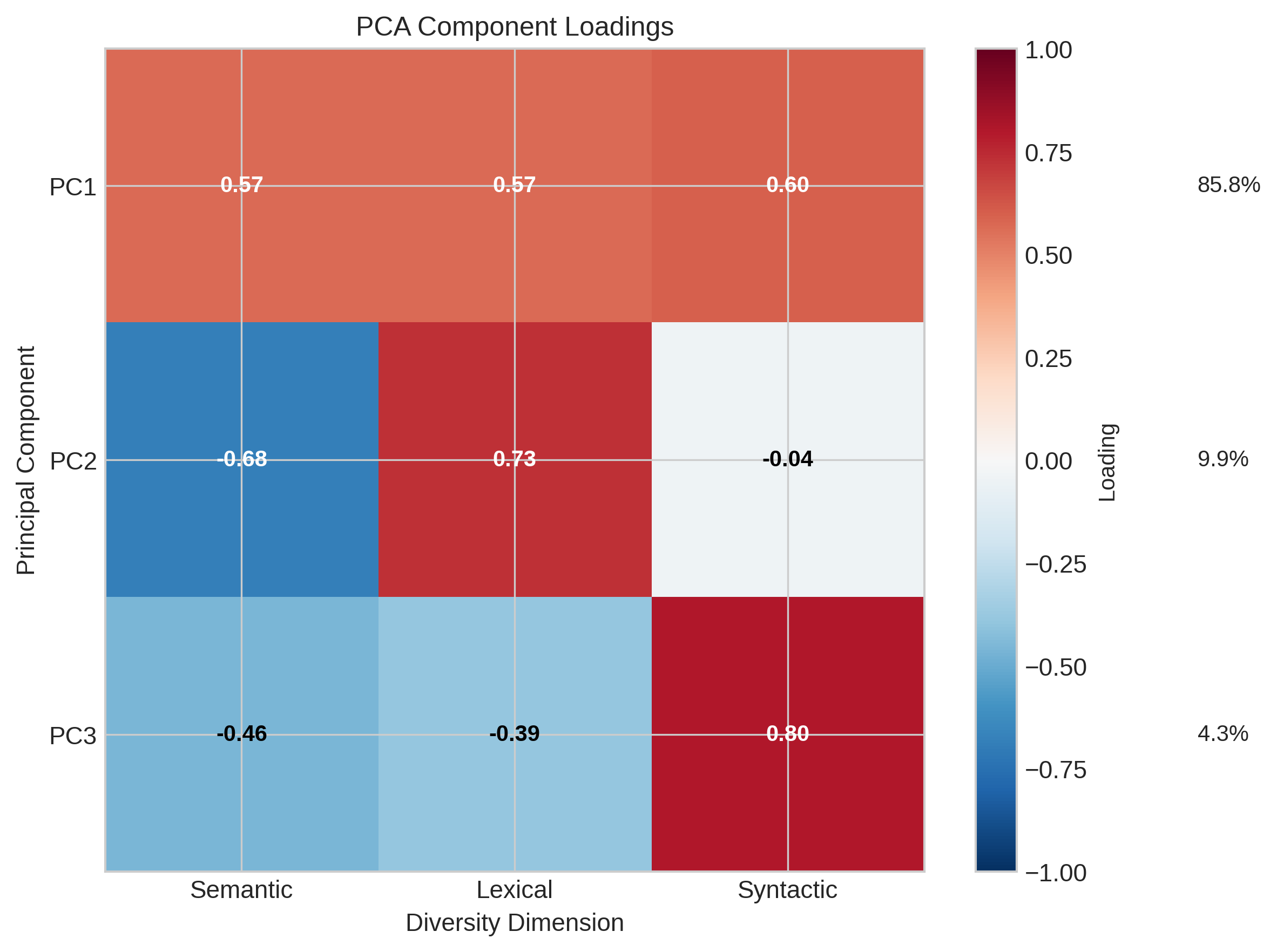}
\caption{Overview of PCA loadings, displaying a balanced contribution of semantic, lexical, and syntactic patterns on the first principal component, which explains a large proportion of the total variance.}
\label{fig:pca_loadings}
\end{figure}

\begin{figure*}[t]
\centering
\includegraphics[width=0.8\linewidth]{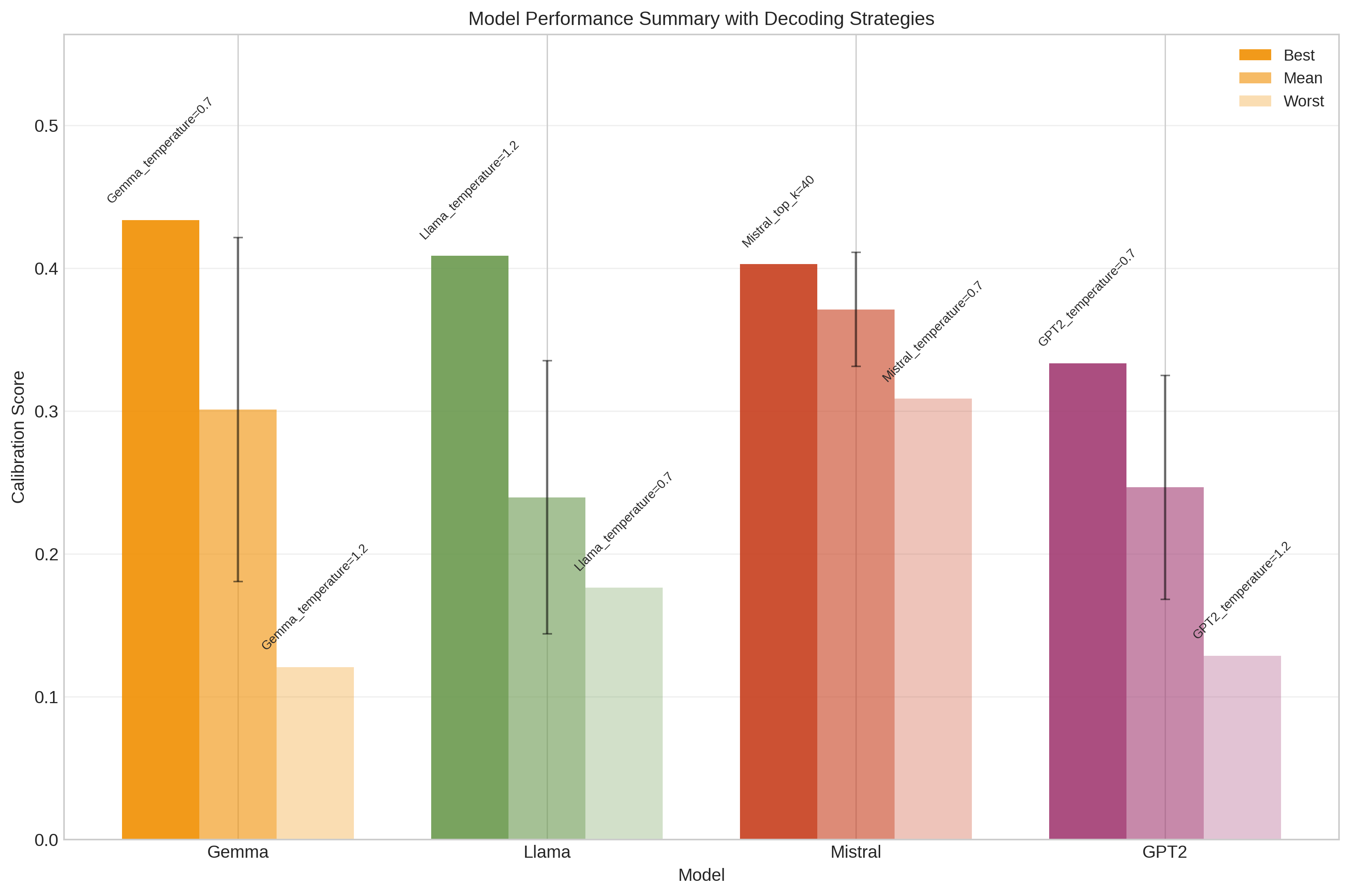}
\caption{Overview of model performance across varying decoding strategies. Here, performance is to be understood in terms of calibration scores with respect to human credal sets. Top-$k$ sampling provides the highest mean calibration, while Gemma-2B with temperature set to 0.7 achieves the best overall calibration.}
\label{fig:uncertainty_decomp}
\end{figure*}

\subsection{Human Variation as Calibration Baseline}

Analysis of 5,000 human-written stories reveals structured patterns of creative variation that establish our calibration baseline (Table \ref{tab:human_baseline}).

\begin{table}[t]
\centering
\small
\begin{tabular}{lcc}
\toprule
\textbf{Diversity Type} & \textbf{Mean} & \textbf{Std Dev} \\
\midrule
Semantic & 0.645 & 0.066 \\
Lexical & 0.328 & 0.035 \\
Syntactic & 0.315 & 0.044 \\
\bottomrule
\end{tabular}
\caption{Human diversity baselines across 500 prompts with 10 unique continuations each, computed using pairwise distance metrics.}
\label{tab:human_baseline}
\end{table}

\noindent The distribution of semantic diversity across prompts shows moderate variation with most prompts (62\%) eliciting medium diversity (0.6-0.7), while 19\% show high diversity (>0.7) and 19\% show low diversity (<0.6). This suggests fundamental differences in prompt interpretability that models must capture.

\subsection{Credal Set Geometry and Calibration}

The human credal set $\mathcal{C}_H$ occupies a volume of 2.25 in the PCA-transformed diversity space, serving as the baseline for model comparison. Analysis reveals a clear distinction between model types: base models (GPT2-XL, Gemma-2B) produce compact credal sets with mean volume $1.10 \pm 0.56$, representing 48.9\% of the human volume. In contrast, instruction-tuned models (Mistral-7B-Instruct, Llama-3.1-8B-Instruct) generate significantly larger credal sets with mean volume $3.87 \pm 1.78$, corresponding to 172.1\% of the human baseline. The difference in credal set volumes between base and instruction-tuned models is statistically significant (Mann-Whitney U = 2.00, $p < 0.001$).

\noindent Principal component analysis of the diversity vectors reveals strong coupling between diversity dimensions. PC1 explains 85.8\% of variance with nearly equal positive loadings across semantic (0.569), lexical (0.565), and syntactic (0.597) dimensions, indicating that these diversity types covary systematically. The dominance of PC1 suggests that models exhibiting high diversity in one dimension tend to show proportionally high diversity in all dimensions, as illustrated in Figure~\ref{fig:pca_loadings}.

\noindent The expanded credal sets of instruction-tuned models indicate broader exploration of the diversity space compared to base models. However, larger volume does not directly correspond to better calibration, as shown in Table~\ref{tab:calibration_results} and Figure \ref{fig:volume_analysis}, in the Appendix. This suggests that alignment with human diversity patterns depends more on the location and shape of the credal set than its absolute size.

\subsection{Distributional Analysis via Wasserstein Distance}

Complementary analysis using Wasserstein distance at the prompt level corroborates the credal set findings. The Wasserstein distance measures the average distributional difference between human and model-generated diversity patterns across all prompts. The best configuration by Wasserstein distance (Gemma-2B with temperature=0.7, distance=0.065) coincides with the best-calibrated credal set, providing independent validation of the geometric approach. The moderate negative correlation between Wasserstein distance and calibration score ($\rho = -0.411$, $p = 0.072$) indicates that while both methods capture aspects of human-model alignment, they emphasize different characteristics: Wasserstein distance weights all prompts equally in measuring average distributional differences, while credal sets capture the geometric envelope of diversity behaviors. A visualization of this comparison is presented in Figure~\ref{fig:wasserstein}.

\begin{table*}[t]
\centering
\resizebox{0.8\linewidth}{!}{
\begin{tabular}{llccccc}
\toprule
\textbf{Model} & \textbf{Strategy} & \textbf{Value} & \textbf{Overall Cal.} & \textbf{Overlap} & \textbf{Centroid Dist.} & \textbf{Volume Ratio} \\
\midrule
Gemma-2B & temperature & 0.7 & \textbf{0.434} & 0.033 & 1.096 & 0.924 \\
Llama-3.1-8B-Instruct & temperature & 1.2 & 0.409 & 0.032 & 1.488 & 0.918 \\
Mistral-7B-Instruct & top\_k & 40 & 0.403 & 0.000 & 1.502 & 1.060  \\
Mistral-7B-Instruct & temperature & 1.2 & 0.399 & 0.000 & 0.956 & 0.820  \\
Mistral-7B-Instruct & top\_p & 0.9 & 0.391 & 0.000 & 1.721 & 1.070  \\
\midrule
Gemma-2B & top\_k & 40 & 0.386 & 0.033 & 1.189 & 0.785 \\
Mistral-7B-Instruct & typical & 0.95 & 0.354 & 0.000 & 1.604 & 1.258  \\
GPT2-XL & temperature & 0.7 & 0.333 & 0.000 & 1.386 & 0.692  \\
Mistral-7B-Instruct & temperature & 0.7 & 0.309 & 0.000 & 1.945 & 1.450  \\
GPT2-XL & top\_k & 40 & 0.300 & 0.033 & 1.244 & 0.509  \\
\bottomrule
\end{tabular}
}
\caption{Calibration metrics for top configurations. Higher values indicate better alignment with human variation. Gemma-2B with temperature 0.7 achieves best overall calibration (0.434).}
\label{tab:calibration_results}
\end{table*}

\subsection{Model Calibration Patterns}

\begin{figure*}[t]
\centering
\includegraphics[width=1\linewidth]{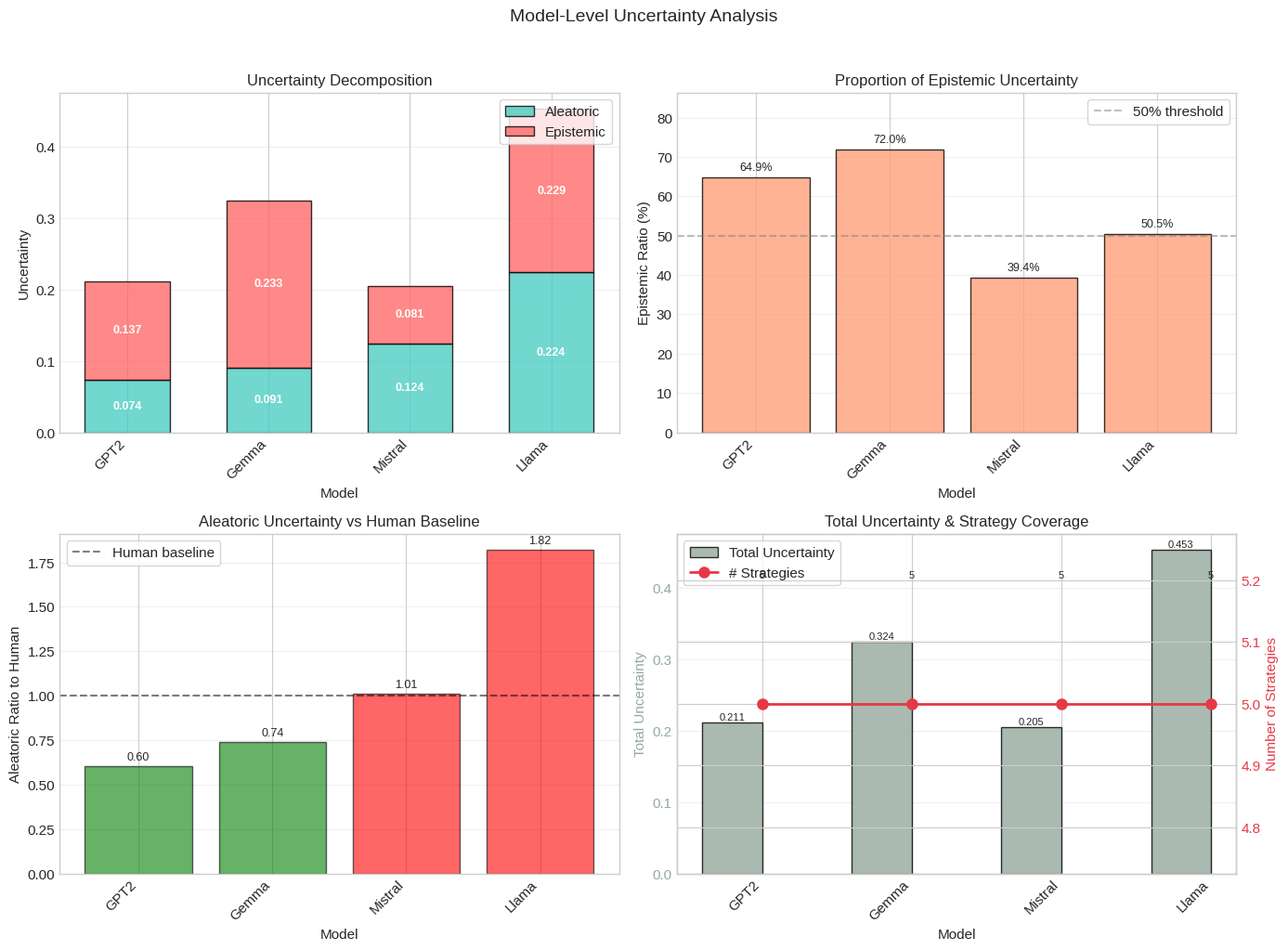}
\caption{Uncertainty analysis and model performance overview. \textbf{Top Left:} Uncertainty decomposition showing epistemic and aleatoric components. \textbf{Top Right:} Epistemic ratio by model. \textbf{Bottom Left:} Aleatoric uncertainty vs. human baseline. \textbf{Bottom Right:} Estimated total uncertainty per model, measured over five decoding strategies.}
\label{fig:calibration_heatmap}
\end{figure*}

Calibration analysis reveals that no model effectively reproduces human variation patterns, with best overlap coefficients reaching only 0.434 (Table \ref{tab:calibration_results}). Figure \ref{fig:calibration_heatmap} illustrates these key findings:

\paragraph{Model architecture effects:} Gemma-2B achieves the best single configuration calibration (0.434 with temperature 0.7), though Mistral-7B-Instruct shows the highest average calibration across all strategies (0.371). Statistical analysis reveals weak positive correlation between model size and calibration (Spearman's $\rho = 0.400$, $p = 0.600$), suggesting model scale has limited influence on calibration quality. Further, base models (mean calibration: 0.274 ± 0.095) show no significant difference from instruction-tuned models (mean: 0.305 ± 0.093) in calibration quality ($t = -0.712$, $p = 0.486$, Cohen's $d = -0.336$). Despite similar calibration scores, base and instruction-tuned models differ significantly in their exploration of the diversity space, with instruction-tuned models producing credal sets 3.5× larger on average ($p < 0.001$).

\paragraph{Strategy effectiveness:} Top-$k$ sampling achieves the highest mean performance (0.323 ± 0.092), followed by temperature scaling (0.289 ± 0.129). Analysis of variance across all 20 model-strategy combinations reveals no significant main effect of strategy type ($F(3,16) = 0.200$, $p = 0.895$), suggesting that strategy effectiveness depends on the specific model architecture.

\subsection{Uncertainty Decomposition}

Decomposition analysis reveals the relative contributions of epistemic and aleatoric uncertainty (Table \ref{tab:uncertainty_decomp}). Base models (GPT2-XL, Gemma-2B) exhibit higher epistemic ratios (64.9-72.0\%), indicating that decoding strategy choice contributes more than half of their total uncertainty. Instruction-tuned models show lower epistemic ratios (39.4-50.5\%), suggesting more stable behavior across decoding strategies but potentially at the cost of reduced overall variation.

\noindent The within-strategy variance (aleatoric component) remains substantial across all models (0.091-0.224), confirming that models can generate diverse outputs for individual prompts. However, the between-strategy variance (epistemic component) highlights that generation configuration remains a critical factor in uncertainty quantification, particularly for base models.

\begin{table}[t]
\centering
\resizebox{\columnwidth}{!}{
\begin{tabular}{lcccc}
\toprule
\textbf{Model} & \textbf{Epistemic} & \textbf{Aleatoric} & \textbf{Total} & \textbf{Ratio} \\
\midrule
Gemma-2B & 0.233 & 0.091 & 0.324 & 72.0\% \\
GPT2-XL & 0.137 & 0.074 & 0.211 & 64.9\% \\
Llama-3.1-8B-Instruct & 0.229 & 0.224 & 0.453 & 50.5\% \\
Mistral-7B-Instruct & 0.081 & 0.124 & 0.205 & 39.4\% \\
\bottomrule
\end{tabular}
}
\caption{Uncertainty decomposition showing absolute values and epistemic ratios. All models show substantial epistemic uncertainty, indicating sensitivity to decoding strategies.}
\label{tab:uncertainty_decomp}
\end{table}

\section{Discussion}

\subsection{Theoretical Implications}

Our credal set framework advances uncertainty quantification theory for generative models in several ways. By treating uncertainty as inherently distributional and prompt-dependent, we move beyond scalar measures that collapse rich variation patterns. The geometric interpretation through credal set operations provides intuitive understanding of miscalibration modes: models can fail through incorrect positioning (wrong variation type), volume (over/under-exploration), or shape (wrong dimensions).

The finding that best calibration reaches only 0.434 reveals fundamental gaps in how current models capture human creative variation. The notably low overlap coefficients (maximum 0.033) indicate that model and human credal sets occupy largely disjoint regions in diversity space, suggesting that current models operate in fundamentally different creative regimes than humans. The high PC1 dominance (85.8\% variance) with syntactic diversity as the primary driver indicates that current models treat diversity dimensions as tightly coupled, potentially missing independent variation patterns that humans explore.

\subsection{Implications for Model Development}

The weak positive correlation between model scale and calibration quality ($\rho = 0.400$, $p = 0.600$) suggests that while larger models may have slight advantages, scale alone is not a determining factor for calibration quality. Our results indicate that training objectives and data distributions likely matter more than parameter count for uncertainty calibration. The lack of significant difference between base and instruction-tuned models ($t = -0.712$, $p = 0.486$, Cohen's $d = -0.336$) with a small effect size indicates that alignment training has minimal impact on creative diversity calibration. Interestingly, instruction-tuned models showed slightly higher mean calibration (0.305 vs 0.274), though this difference was not statistically significant. The substantial epistemic uncertainty across all models (39.4-72.0\%) highlights that decoding strategy choice remains a dominant source of variation. Notably, Gemma-2B shows the highest epistemic ratio (72.0\%), suggesting extreme sensitivity to decoding configuration despite achieving the best single-configuration performance. This paradox suggests that optimal calibration may require careful strategy selection rather than robust performance across strategies.

\subsection{Practical Deployment Considerations}

For practitioners deploying language models in creative applications, our findings offer concrete guidance:

\begin{itemize}
    \item \textbf{Model selection}: Mistral-7B-Instruct offers the most consistent performance across strategies (mean calibration: 0.371), while Gemma-2B with temperature 0.7 provides the best single configuration (0.434).
    
    \item \textbf{Strategy optimization}: Top-$k$ sampling provides the highest mean calibration (0.323), though all models show substantial epistemic uncertainty (39-72\%), making careful tuning essential.
    
    \item \textbf{Baseline expectations}: With maximum calibration at 0.434 and overlap coefficients of at most 0.033, expect substantial divergence from human creative patterns.
    
    \item \textbf{Multi-strategy ensemble}: Given high epistemic ratios, combining outputs from multiple decoding strategies is crucial for approximating human creative diversity.
    
    \item \textbf{Model-specific tuning}: In terms of calibration, base models (especially Gemma-2B at 72\% epistemic) require more careful strategy selection than instruction-tuned models like Mistral-7B-Instruct (39.4\% epistemic).

    \item \textbf{Calibration vs. quality}: Calibration along semantic, lexical, and syntactic dimensions does not necessarily indicate qualitative alignment between model-generated and human-produced text. Future work will investigate this relationship comprehensively using both human evaluations and LLM-as-a-Judge scores.

    \item \textbf{Generalizatbility}: Our findings are specific to \textit{storytelling}—an open-ended task prioritizing communicative goals such as creativity, fluency, and engagement. To extend this analysis to other Natural Language Generation (NLG) research areas, we suggest task-specific calibration analyses, as different tasks involve distinct communicative objectives and varying degrees of human production variability that serve as calibration benchmarks.

\end{itemize}

\section{Conclusion}

This work establishes credal sets as a rigorous framework for uncertainty quantification in open-ended text generation, enabling principled geometric comparison between human and model variation patterns. Through comprehensive analysis of 100,000 generated stories calibrated against 5,000 human-written stories, we demonstrate substantial gaps in how current language models capture human creative variation, with the best calibration reaching only 0.434 (Gemma-2B with temperature 0.7) and overlap coefficients at most 0.033.

\noindent Our decomposition reveals that epistemic uncertainty from decoding strategy choice contributes 39.4-72.0\% of total uncertainty across models, with base models showing higher sensitivity to configuration choices. The weak correlation between model scale and calibration ($\rho = 0.400$, $p = 0.600$) and lack of significant difference between base and instruction-tuned models ($p = 0.486$) challenge common assumptions about model development priorities. The credal set framework provides actionable insights for deploying language models in creative contexts and establishes quantitative benchmarks for evaluating progress toward human-AI creative alignment. As language models increasingly engage in open-ended generation tasks, our findings highlight the critical importance of decoding strategy selection and the need for architectural or training innovations specifically targeting uncertainty calibration. 

\section*{Limitations}

Several limitations warrant consideration:

\begin{itemize}
    \item Our analysis uses convex hulls which may not capture non-convex uncertainty regions or multimodal distributions within credal sets.
    \item The 500-prompt sample from a single domain may not generalize to other creative writing contexts or  languages.
    \item Decoding strategies evaluated prioritize high-probability tokens, whereas humans often select surprising, low-probability tokens for creative effect—a mismatch that may constrain achievable calibration.
    \item Human baselines include natural skill variation beyond pure creativity, potentially inflating aleatoric uncertainty estimates.
    \item Computational constraints limited us to 10 samples per configuration; larger samples might reveal finer-grained patterns.
    \item Statistical variance alone cannot distinguish creative quality from random variation—validating the relationship between our metrics and perceived creative quality is essential future work.
\end{itemize}

\section*{Ethics Statement}

We affirm that our research adheres to the \href{https://www.aclweb.org/portal/content/acl-code-ethics}{ACL Ethics Policy}. This work uses publicly available datasets and involves no human subjects or personally identifiable information. We acknowledge potential biases in the Reddit-sourced dataset and encourage diverse dataset development. Our framework could help identify and mitigate generation biases by comparing model variation patterns across different demographic or cultural contexts. All code and data are released to enable reproducible research and further investigation of these important issues.

\section*{Acknowledgments}

Esteban Garces is sponsored by the Munich Center for Machine Learning (MCML) and the LMU Mentoring Program. Julian Rodemann acknowledges support by the Bavarian Institute for Digital Transformation (bidt) within the Bavarian Academy of Sciences (BAS) and the LMU Mentoring Program. 

\bibliography{custom}

\clearpage

\onecolumn

\appendix

\section{Extended Results}
\label{app:extended_results}

\subsection{Credal Volume Analysis}

\begin{figure*}[ht!]
\centering
\includegraphics[width=1\linewidth]{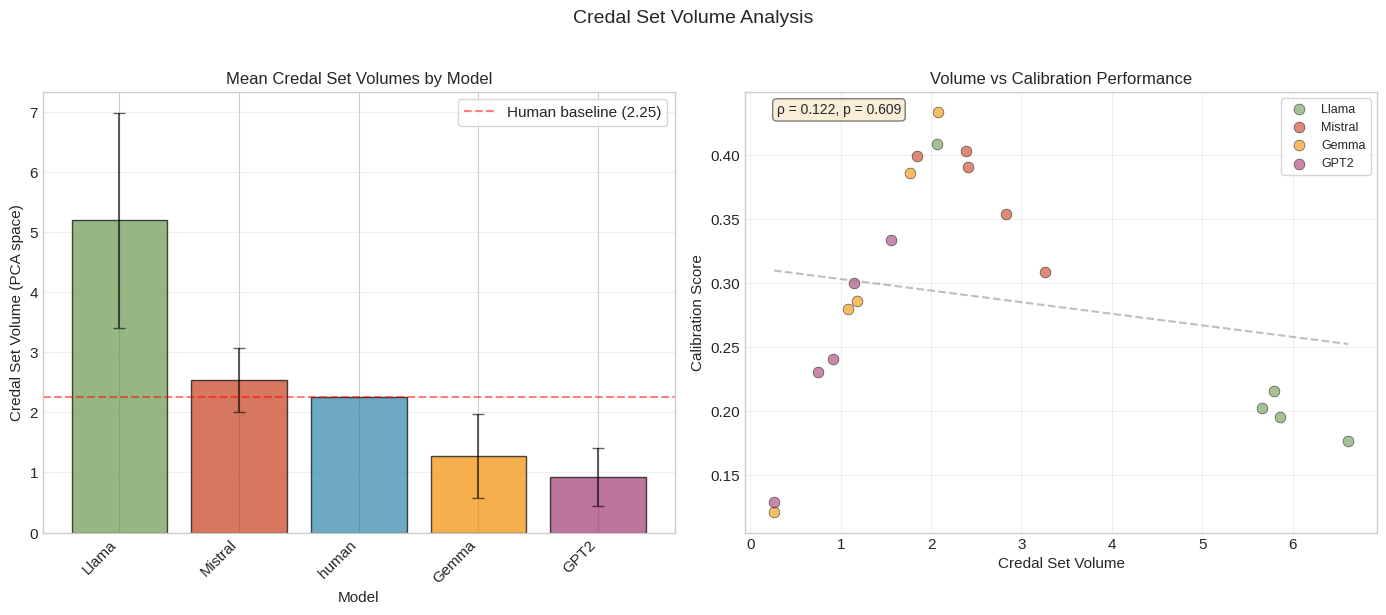}
\caption{Analysis of credal set volumes for human and language models. \textbf{Left:} Mean credal set volumes by model (in PCA space). \textbf{Right:} Relationship between calibration score and credal set volume. A positive trend for base models (GPT2-XL and Gemma) is observed, while a negative trend is observed for instruct models (Mistral and Llama).}
\label{fig:volume_analysis}
\end{figure*}

\subsection{Wasserstein Distance Analysis}

\begin{figure*}[ht!]
\centering
\includegraphics[width=1\linewidth]{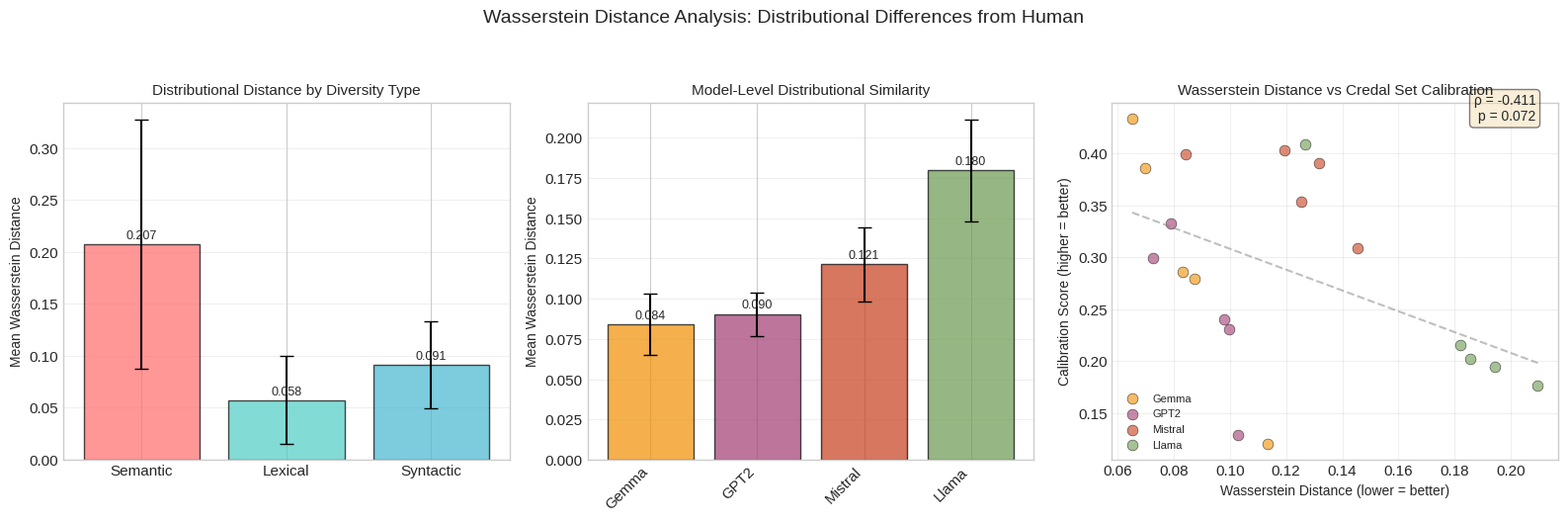}
\caption{Distributional differences between model and human productions measured by Wasserstein distances. \textbf{Left:} Mean Wasserstein distances across semantic, lexical, and syntactic dimensions. Semantic features show the largest divergence from human distributions, followed by syntactic and lexical features. \textbf{Middle:} Model-specific distributional similarity. Gemma-2B achieves the lowest Wasserstein distances (closest to human distributions), while Llama models exhibit the highest distances. \textbf{Right:} Inverse relationship between calibration scores and Wasserstein distances (moderate negative correlation). Gemma-2B and Mistral appear in the upper-left section (high calibration, low distance), while Llama appears in the lower-right quadrant (low calibration, high distance).}
\label{fig:wasserstein}
\end{figure*}

\clearpage

\subsection{Complete Calibration Results}

Table \ref{tab:full_calibration} presents calibration coefficients for all 20 model-strategy combinations evaluated in our experiments.

\begin{table}[h]
\centering
\small
\begin{tabular}{llc}
\toprule
\textbf{Model} & \textbf{Strategy} & \textbf{Calibration} \\
\midrule
Gemma-2B & temperature\_0.7 & 0.434 \\
Llama-3.1-8B-Instruct & temperature\_1.2 & 0.409 \\
Mistral-7B-Instruct & top\_k\_40 & 0.403 \\
Mistral-7B-Instruct & temperature\_1.2 & 0.399 \\
Mistral-7B-Instruct & top\_p\_0.9 & 0.391 \\
Gemma-2B & top\_k\_40 & 0.386 \\
Mistral-7B-Instruct & typical\_0.95 & 0.354 \\
GPT2-XL & temperature\_0.7 & 0.333 \\
Mistral-7B-Instruct & temperature\_0.7 & 0.309 \\
GPT2-XL & top\_k\_40 & 0.300 \\
Gemma-2B & top\_p\_0.9 & 0.286 \\
Gemma-2B & typical\_0.95 & 0.279 \\
GPT2-XL & top\_p\_0.9 & 0.240 \\
GPT2-XL & typical\_0.95 & 0.231 \\
Llama-3.1-8B-Instruct & typical\_0.95 & 0.215 \\
Gemma-2B & temperature\_1.2 & 0.212 \\
Llama-3.1-8B-Instruct & top\_k\_40 & 0.199 \\
Llama-3.1-8B-Instruct & temperature\_0.7 & 0.196 \\
GPT2-XL & temperature\_1.2 & 0.188 \\
Llama-3.1-8B-Instruct & top\_p\_0.9 & 0.175 \\
\bottomrule
\end{tabular}
\caption{Complete calibration results for all model-strategy combinations, sorted by calibration coefficient.}
\label{tab:full_calibration}
\end{table}

\subsection{Statistical Tests}

We conducted comprehensive statistical analyses to validate our findings:

\begin{itemize}
\item \textbf{Model size vs calibration}: Spearman's $\rho = 0.400$ ($p = 0.600$), indicating weak positive correlation without statistical significance.
\item \textbf{Base vs instruction-tuned}: Two-sample t-test: $t = -0.712$ ($p = 0.486$), no significant difference. Cohen's $d = -0.336$ (small effect size).
\item \textbf{Strategy comparison}: ANOVA across strategies: $F(3,16) = 0.200$ ($p = 0.895$), no significant differences between strategies.
\item \textbf{Best performing model}: Mistral-7B-Instruct showed highest mean calibration (0.371) across all strategies.
\item \textbf{Best performing strategy}: Top-$k$ sampling achieved highest mean calibration (0.323 ± 0.092) across all models.
\end{itemize}

\section{Implementation Details}
\label{app:implementation}

\subsection{Computational Resources}

All experiments were conducted on Google Colab with the following specifications:
\begin{itemize}
\item GPU: NVIDIA A100 (40GB) or V100 (16GB)
\item RAM: 25-50GB depending on instance
\item Storage: Google Drive for persistent storage
\item Total compute time: Approximately 8 hours for generation, 1 hour for analysis
\end{itemize}

\subsection{Model Configurations}

Models were loaded with the following optimizations:
\begin{itemize}
\item 4-bit quantization for models >3B parameters using BitsAndBytes
\item Flash Attention 2 where supported
\item Batch sizes optimized per model (8-25 samples)
\item Automatic mixed precision (AMP) with fp16
\end{itemize}

\subsection{Diversity Metric Computation}

Semantic embeddings were computed using Sentence-BERT (all-MiniLM-L6-v2) with the following parameters:
\begin{itemize}
\item Maximum sequence length: 512 tokens
\item Batch size: 64 for encoding
\item Pooling: Mean pooling over token embeddings
\end{itemize}

POS tagging was performed using spaCy's en\_core\_web\_sm model with a maximum text length of 5000 characters for efficiency.

\end{document}